# Line Segmentation from Unconstrained Handwritten Text Images using Adaptive Approach

Nidhi Gupta[1], Wenju Liu[2]

Mathematics and Scientific Computing, National Institute of Technology, Hamirpur, Himachal Pradesh, India

National Laboratory of Pattern Recognition, Institute of Automation, Chinese Academy of Sciences, Beijing, China

nidhi@nith.ac.in, lwj@nlpr.ia.ac.cn

**Abstract** Line segmentation from handwritten text images is one of the challenging task due to diversity and unknown variations as undefined spaces, styles, orientations, stroke heights, overlapping, and alignments. Though abundant researches, there is a need of improvement to achieve robustness and higher segmentation rates. In the present work, an adaptive approach is used for the line segmentation from handwritten text images merging the alignment of connected component coordinates and text height. The mathematical justification is provided for measuring the text height respective to the image size. The novelty of the work lies in the text height calculation dynamically. The experiments are tested on the dataset provided by the Chinese company for the project. The proposed scheme is tested on two different type of datasets; document pages having base lines and plain pages. Dataset is highly complex and consists of abundant and uncommon variations in handwriting patterns. The performance of the proposed method is tested on our datasets as well as benchmark datasets, namely *IAM* and *ICDAR*09 to achieve 98.01% detection rate on average. The performance is examined on the above said datasets to observe 91.99% and 96% detection rates, respectively.

## 1. Introduction

In today's era of digitalized world, most of the information is preferably stored in the digital format. However still sometimes, answer sheets, official tasks, bank transaction slips, degrees and many other important documents are on paper work and hence considered to be more reliable. Needless to mention, that importance of the papers cannot be underestimated even in the era of digitalized world. Still on business world, it is cheaper and more practical to share information rather than electronic means. In one survey it is observed that over 95% of official data are still on papers and could not be digitalized. People can easily handle such information on paper and easily convert into a suitable format for further processing. Optical character recognition is one of the researches in document analysis field, where words and characters are to be recognized automatically to reduce the manual error and undesired extra efforts. One can easily realize the importance of the line segmentation and word recognition task in document image analysis. For this, line segmentation is one major task to be handled before word or letter recognition. But line segmentation from handwritten images is a major challenging task for character recognition models. The raw document text images have several kinds of diversity and unknown variations. In the literature, numerous works are accomplished and aimed for the line segmentation, but either their models are dominant to specific dataset or inclined to fixed inherent parameters. It can be stated well that the provided results are limited up to some extent [14] [2].

Based on above discussed limitations, the objective of the proposed work is to develop robust and adaptive technique towards line segmentation. The robustness deals with any kind of handwriting styles, whereas adaptiveness deals for successful execution of algorithm irrespective to dataset type or size. Segmentation methods are broadly divided into two categories; empirical methods and analytical methods. Empirical methods work indirectly on test images and produce results, whereas analytical methods directly tested on images and analyze the properties. Before such kind of implementation, sometimes raw handwritten images need to be processed. Moreover it is a good choice to pre-process raw data in order to deal with only informative and significant pixels, despite processing on whole document image. Thresholding, cropping, sharpening or enhancing are a few listed conventional and well known pre-processing techniques. After complete successful application of such methods, post processing has also been implied in some methods to rectify irrelevant and insignificant segmented lines [2] [8].

Further the methods used for the line segmentation is one of the most critical task for the accurate segmentation. Most of the existing methods are non-robust and infeasible to implement. These methods work excellent under certain conditions or prerequisites. Therefore, the generalized scheme is yet not achieved by any existent system mentioned in the related works.

In this present work, a new scheme is proposed using connected component coordinates. The coordinates are matched to find the text into aligned margin. The novelty of the work lies in finding the text height dynamically. In the literature, it is been proven as a hurdle to find a text height dynamically, more specifically when texts are located at unconstrained and unpredictable location. A mathematical formula is derived to determine the text height respective to the image size using statistical proof. Further the coordinates and text height are matched to segment lines. The major advantage of the work is to implement and efficiently work on our composed dataset as well publicly available datasets *IAM* and *ICDAR09*, both respectively.

The organization of the manuscript is as follow. Section 2 includes literature review and associated shortcomings throughout. Section 3 describes the information of benchmark dataset as well our experimental dataset which is used to perform the proposed scheme. The proposed methodology is given in Section 4 with each step in detail. Section 5 consists of results and discussion of the proposed scheme. Comparisons with other existing methods and performance of other methods on our datasets are given in Section 6. At last, Section 7 concludes the work.

## 2. Related Work

Most of the related existing works can be broadly summarized into major categories, like smearing approach, transformation approach, profile projection, active contour algorithms and methods based on morphological properties. However, such techniques are bound to process under certain conditions and cannot be used directly or independently. Among transformation-based approaches Hough transformation is widely used to deal with slanting and oriented lines. Other techniques are not promising to provide an efficient segmentation. Many of the methods are based on modifications on conventional and trivial techniques. A few attractive and well known methods are discussed here in brief.

In 2004, Shi [13] proposed a line separation method from handwritten document image using fuzzy directional run length. The image is converted into fuzzy image and run lengths are calculated from binarized fuzzy image. Connected components are figured out and pattern of text lines are estimated. The algorithm is tested on postal parcel images and historical handwritten documents such as Newton's and Galileo's manuscripts to achieve 93% rate on overall performance.

In 2009, Louloudis [3] proposed a scheme using Hough transform on the subset of document image using connected components. A post-processing step includes the correction of possible false alarms, the detection of text lines that Hough transformation failed to create. Afterwards, for the separation of vertically connected characters the skeletonization method is used. The novelty lies in the improvement of such kind of separation of vertically connected text lines. The distance metrics comprised of the Euclidean distance metric and the convex hull-based metric. Two class problems are considered as an unsupervised clustering problem and Gaussian mixture theory is used for the solution. The performance of the scheme is examined on a consistent and concrete evaluation methodology. They demonstrated results on two datasets namely *ICDAR2007* and a historical dataset.

In 2010, Papavassiliou [5] proposed a line segmentation algorithm using Viterbi algorithm based on locating the optimal succession of text and gap areas within vertical zones. The document image is divided in vertical zones and the extreme points of the piece-wise projection profiles are used to over-segment each zone in gap and text regions. Then, statistics of a Hidden Markov Model are estimated to feed Viterbi algorithm in order to find the optimal succession of text and gap are as in each zone. Line separators of adjacent zones are combined with respect to their proximity and the local foreground density. Text line separators are drawn in staircase function fashion across the whole document. Finally, connected components are assigned to text lines or are cut in a suitable junction point of their skeleton by applying a simple geometrical constrain. The algorithms tested on the benchmark datasets of *ICDAR07* handwriting segmentation contest.

In 2011, Alaei [1] employed a smearing operation on the foreground portion of the document image. The painting technique enhances the separability between the foreground and background portions to enable text-line detection. A dilation operation is employed on the foreground portion of the painted image to obtain a single component for each text-line. Thinning of the background portion of the dilated image and subsequently some trimming operations are performed to obtain a number of separating lines. Related candidate line separators are connected to obtain the segmented text-lines by using the starting and ending points of the candidate line separators and analysing the distances among them. Furthermore, the problems of overlapping and touching components are addressed. They tested the proposed scheme on text pages of around seven different languages and remarkable results are presented.

Klette in 2015 [17], proposed a statistical method based on the Hough transform to extract the set of parameters of a line segment. Later they enhanced the methodology for line segment detection by considering quantization error, image noise, pixel disturbance, and peak spreading [18]. Calculating statistical parameters, non-zero cells are analyzed and computed followed by interpolation and curve fitting. The experimental results are shown on simulated data and real world images to validate the line segmentation performance.

In 2018 lee [16] developed linelet based representation to model intrinsic properties of line segments in rasterized image space to construct frameworks for line segment detection, validation, and aggregation. They experimented on their own dataset to show comparative performance from state-of-the-art methods.

Again, in 2018, Sen [4] proposed a scheme based on stroke-level busy zone formation procedure. A sub-zoning scheme within busy zone followed by a modified down-up-down concept within these sub-zones has been used to find valid segmentation points. This scheme avoids over and under segmentation issues. These issues are related to either inherent writing pattern or due to writing style variations up to certain extent. The proposed segmentation approach are tested on 6,500 online handwritten Bangla word samples with 98.45% correct segmentation accuracy tested on manually generated ground truths.

Recently, in 2019 Li [6] proposed a text line segmentation method based on two steps; baseline detection and text line segmentation using the baseline. Horizontal gradient operator is applied upper edges of all characters in the document and upper edge set was divided into disjoint subsets using edge connectivity. The eligible subsets further used to obtain the baseline. In second step of text line segmentation, image was truncated at baseline position and adhesion regions are segmented again. The connected regions belonging to baseline formed a text line. Experiments performed on dataset showed that the method is effectively avoiding document distortion and high accuracy achieved with the handle of text line adhesion.

In the same year, in 2019 Dlya [11] proposed a fast and adaptive bi-dimensional empirical mode decomposition method for line segmentation task. The document image is decomposed into two components. The image is binarized using Niblack's and Sauvola's Algorithms. Further, text line segmentation is achieved with the elimination of bi-dimensional intrinsic mode functions. The results show that the method is noise resistible and provide improved version of text line segmentation.

We studied several papers and found similar observation on the provided results. The results are highly dependent and prone to dataset variations. The results shown by the research papers are not robust and fails to achieve accuracy in case of different datasets. Like transformation based methods can only be useful on some handwritten images which are aligned in some angle or slanted towards base line. Smearing based methods are useful only when handwritten text lines are in smeared form. Projection based methods work efficiently but fails when occurs with skewness and if examined for unconstrained images. In order to tackle this problem, first skewness correction is required. Moreover the experimental dataset is another challenging task for verification and validation of such systems for real time applications. Most of the works are applied on small dataset compared to our dataset size. The said limitations are tried to be solved and results are presented to emphasis on mathematical solution for line segmentation. In the proceeding section, the proposed method is elaborated and discussed.

## 3. Experimental Dataset

The paper represents a wide dataset of handwritten images. The raw document handwritten unconstrained text images dataset are composed at local software company located in Beijing, China. These images are of two types; having handwritings on plain images and ruled pages. Both of the dataset images are provided with the ground truth templates. The company aims to produce real time application based products for document image analysis. Dataset contains in total 12,909 images, where 6,517 handwritten images are on plain pages and 6,392 images are handwritten images on ruled pages. Handwritings are in unconstrained nad non-uniformly written. The handwritings are in random order and in unpredictable format. It is further divided into two forms; plain and rule pages. In context for further notation, Dataset I reflects plain handwriting images and Dataset II reflects handwritings written on the ruled pages (with base lines). It is difficult to analyze the pattern of writing styles.

Another publicly available datasets *IAM* and *ICDAR* [7] are used for the uniform evaluation and testing the performance of the proposed method. *IAM* handwriting database contains English handwritten text and used to perform writer identification and verification experiments. It was first published in the year 1999 at *ICDAR* conference and used to hold per year as a robust reading competition. Approximately, it contains around 1300 images with more than three different text orientations, horizontal, multi-oriented, and curved ones.

In *ICDAR09*, the dataset images are manually annotated in order to produce the ground truth, which corresponds to the correct text line and word segmentation result. For the evaluation, a well-established approach is used based on counting the number of matches between the entities detected by the segmentation algorithm and the entities in the ground truth. The encoded ground truth files are given in binary format and hence decoded for the extraction of number of lines for the verification of results.

4. Proposed Method

The paper has represented precise and effect method for segmenting line from unconstrained dataset. The framework of the proposed scheme is shown in Fig. 1 as below.

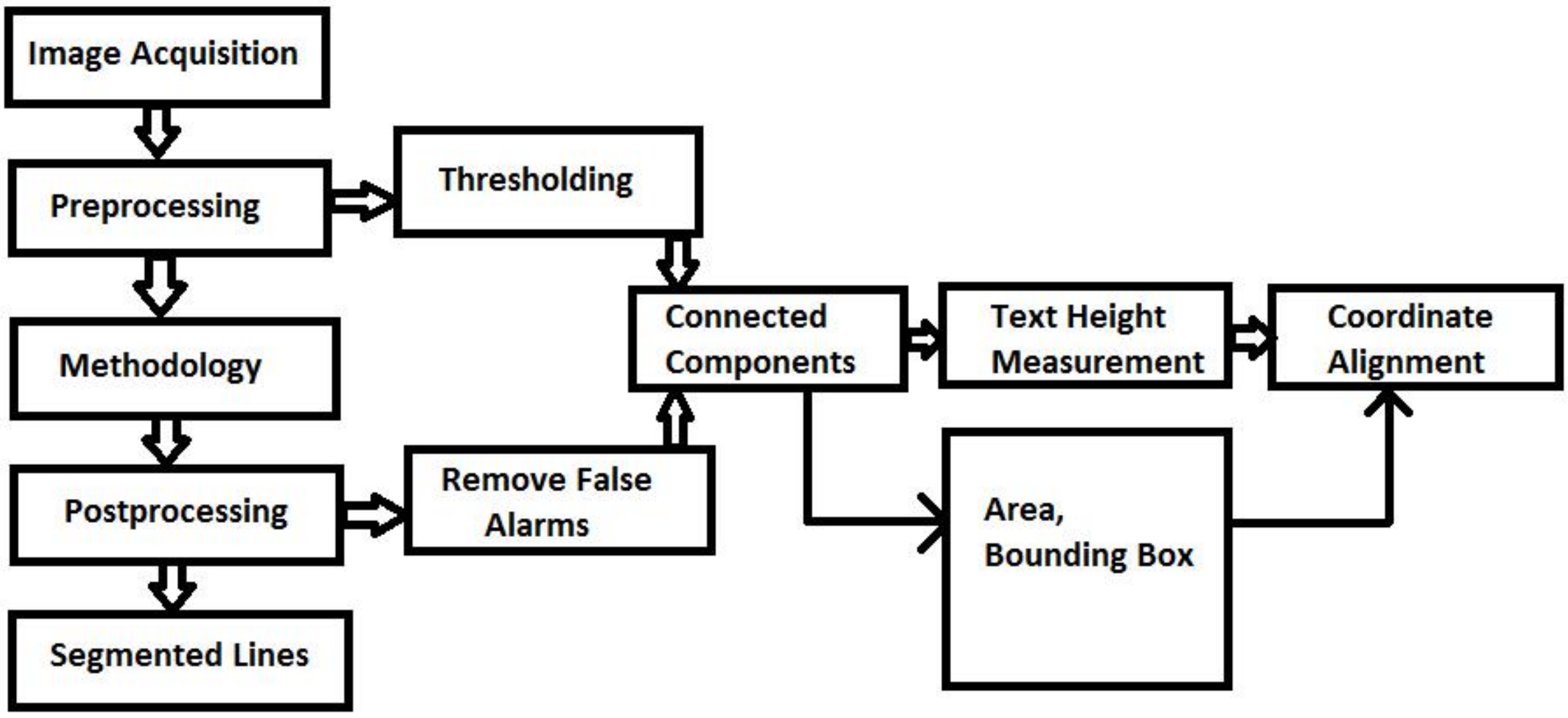

Figure 1: Framework of an adaptive line segmentation scheme from unconstrained text image

Line segmentation task is broadly divided into major steps for accurate line segmentation. Initially, base line is measured through the coordinate alignment of connected components. The connected components are created with specified constraints. Later, text height are measured dynamically respective to the size of the document image. The novelty lies in finding text height with statistical background. The scheme is divided into several steps as described in the following steps.

4.1 Pre processing

Sometimes text document images require pre-processing on raw document images due to blurriness and not readable form. Many of the pre-processing steps include processing methods like, noise reduction, enhancement, sharpening, smoothening, cropping, smoothing, thresholding etc. The applications of such methods are quite feasible and efficient, but operations on pixel original information sometimes lead wrong outcome or distort original information. Therefore it is a requirement to separate significant pixels from the entire image. The optimal threshold is measured
by following formula as,

$$Threshold = Image_max - \sigma$$

where, σ is a standard deviation of the image. Images are thresholded to separate background and foreground pixels. The foreground pixels are thus considered as significant pixels consist of useful text information for accurate line segmentation and processed for the further task.

4.2 Algorithm

This is a major part applied for the line segmentation task from handwritten document images. It is divided into two main sub-steps as discussed below.

4.2.1 Baseline Detection and Coordinates Alignment

The concept of bounding box coordinates is compact together with text height measurement for the adaptation of inherent properties from both. Further, connected components are extracted from eight-pixel neighbourhood from the foreground pixels. The coordinates of connected components are measured as $x, y$ (indices) with four registered associated labels $x_{min}$, $x_{max}$, $y_{min}$, and $y_{max}$. These coordinates are aligned in a manner that is distant by less than fifteen pixels. The method is demonstrated in Fig. 2.

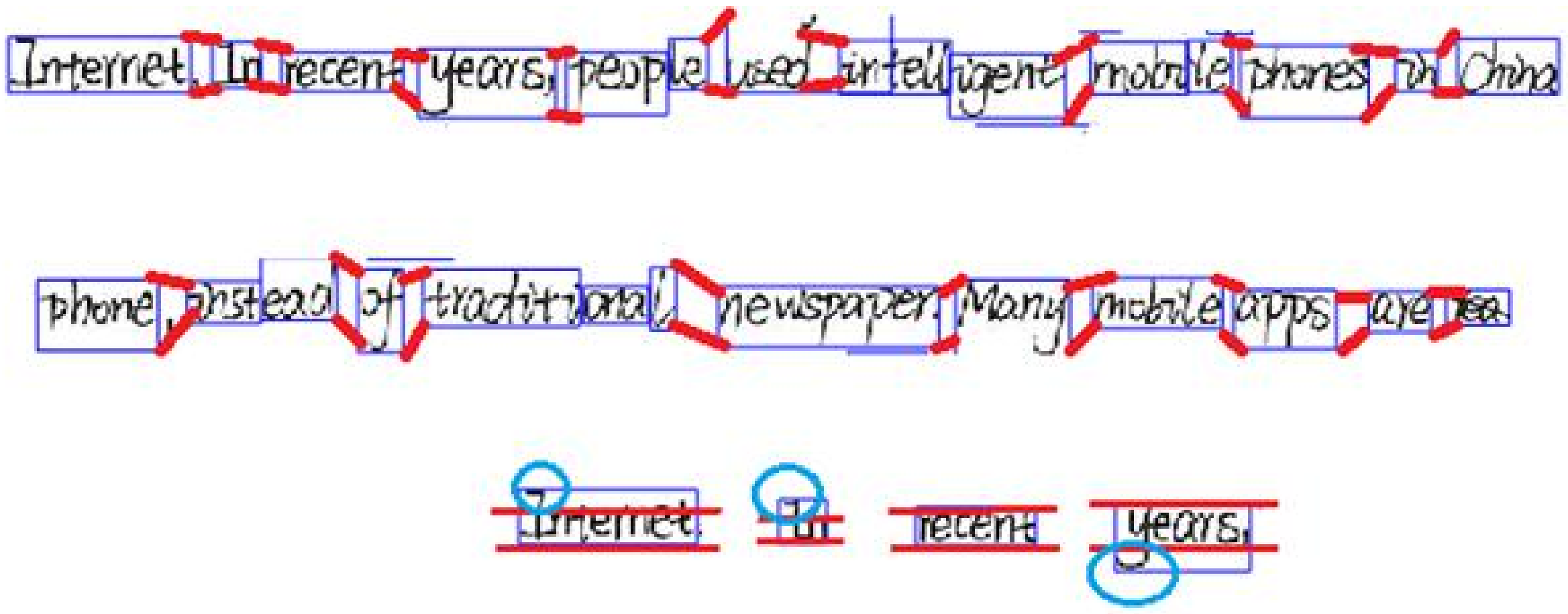

Figure 2: Baseline detection and coordinates alignment

A regional property in a specified range is used to draw the connected components. Further, the coordinates of each connecting components are stored and corresponding two regional features namely, area and bounding box are extracted. Area signifies the numbers of pixels present in the blob and bounding box specifies the boundary coordinates of the selected blob. The blobs are selected by measuring area found in the specific range. Appropriate blob detection is another challenging task for the line segmentation task. Blobs are calculated by counting significant number of pixels present in it. In the proposed scheme, area is taken in significant range and such pixel counts provide substantial effect on the processing. Further, the coordinates of each box are aligned in such a manner that is distant by fifteen pixels at maximum. The characters are usually written in same size except some kind of extra strokes in the characters. In the figure 2, the extra strokes are considered out of the range of fifteen pixels at maximum. Blue circle denotes extra strokes for the words, so the bounding boxes are drawn with different coordinates. The marked pixels are considered to be aligned vertically on one line. In this way, $x_{min}$ and $x_{max}$ are constructed in an array.

### 4.2.2 Text Height Measurement

Considering the characteristics of the text written on images can effectively enhance the detection rate of the line segmentation task. The novelty of the proposed scheme lies in finding the text height dynamically. Height of the text is mathematically driven by information based on the size of the provided input document image. The text height is calculated by the formula as,

$$Height_{text} = \left(\frac{1}{24}\right)\sqrt{\left(\frac{width}{2}\right)^2 + height^2}$$

Statistical Derivation:

The formula is derived with the phenomenon of the input image height and width information. Text height can be measured explicitly with the prior knowledge of size of the image. The measurement of extent boundary coordinates of the text in each line is applied. Statistically, height and width are substitute of information of an image, which provide the probability of handwritings stroke. In the most important step of proposed scheme, both of the collected information is compacted, *i.e.*, bounding box coordinates and text height. The half width is considered to achieve appropriate contents inside the text image. The coordinates of each bounding boxes are aligned and insignificant coordinates are discarded. The upper and lower coordinates define the height of the boxes. The largest gap between upper coordinate and lower coordinate is taken place for the measurement. Therefore, if the coordinates of the bounding boxes and text height are found in the range, then the segmentation falls correctly else discarded. The steps are illustrated in Algorithm 1 as below. Fig. 3 represents the stepwise graphical algorithm.

Algorithm *Line Segmentation from Handwritten Images*

*Input*: Document Text Image
*Output*: Segmented Lines

*begin*

*read an RGB image I of size (m, n, 3)*

$new_{size(m,n)} <== size(m, n, 3)$

$(max(I) - \sigma) > I$

    $I_f <== I$

*else*

    $I_b \leq== I$

$temp <== cc(I_f)$

*measure area and bounding boxes from temp*

*measure height of the text*

$$Height_{text} = (\frac{1}{24})\sqrt{(\frac{width}{2})^2 + height^2}$$

*where height and width are dimensions of temp*

*align the extracted coordinates*

*count white pixels inside the line segment*

*if count > 30%*

    *break()*

    *else*

    *continue()*

*end*

4.3 Post processing

Post processing is applied for the correction of the false alarms before stopping the algorithm. It concerns the use of linguistic constraints to improve segmentation performance. If number of the white pixels are present more than 30% of the total pixels, then segmented lines are discarded. The performance of the line segmentation is checked with matching the number of segmented lines from corresponding template image on ground truth templates. The segmentation is declared successful, if the number matches from the ground truth.

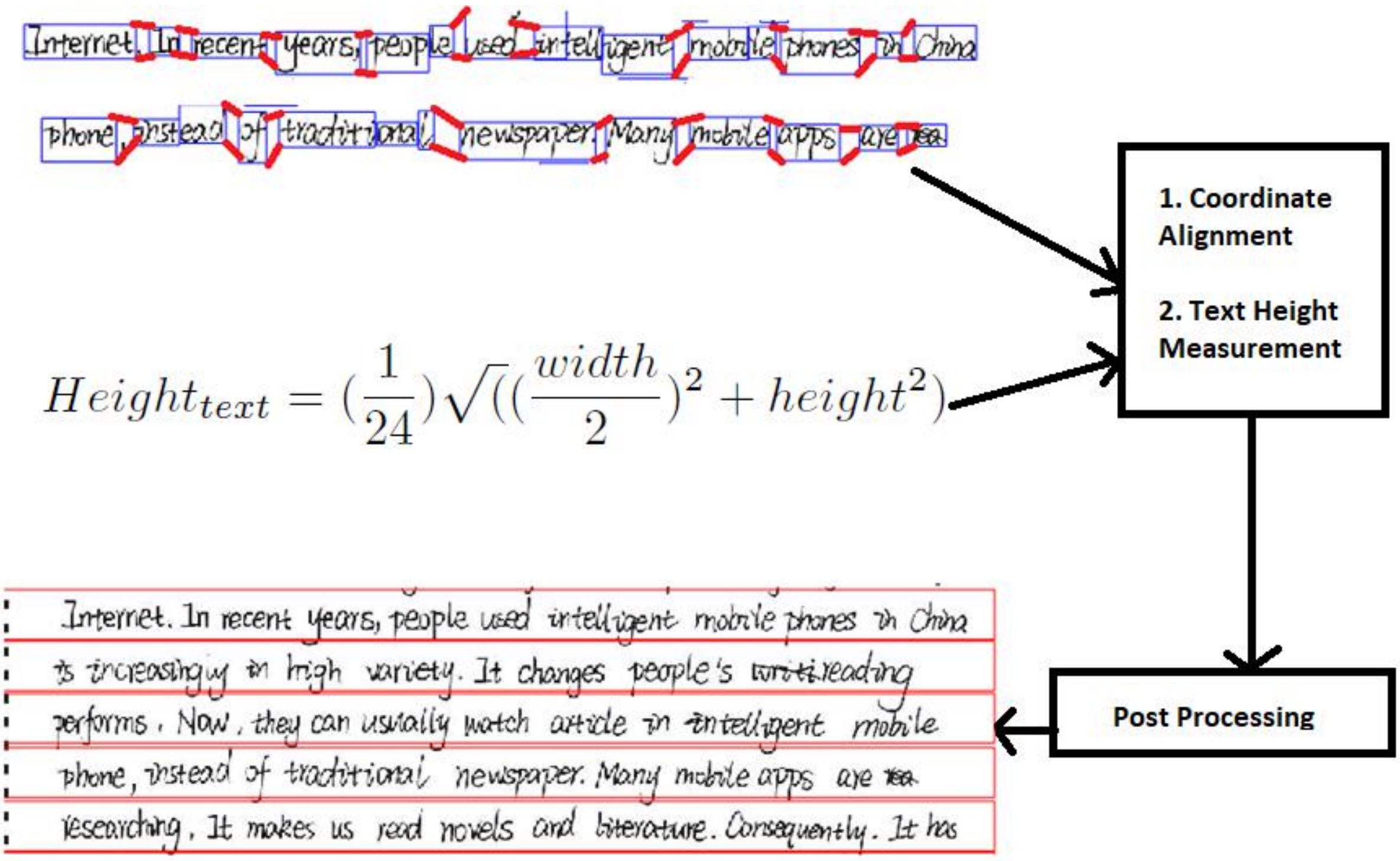

Figure 3: Illustration of the proposed scheme

Fig. 3 depicts the proposed scheme with detailed illustration specifying each step. At first, the bounding boxes limit coordinates are extracted and aligned. Next, text height is calculated through statistically derived formula. Based on these steps, the coordinates are finalized and aligned in the margin. Post processing includes to discard the segmented lines, if consist of white pixels more than 30% of the remaining pixels. The proposed scheme is explicitly described in the flow diagram in details.

## 5. Results and Discussion

The section describes the results and discussion of the proposed scheme. Later, it adds the comparative studies with the other existing methods. The scheme is also cross validated on benchmark datasets to observe the robustness and efficiency.

### 5.1 Experimental Setup

The system for line segmentation from unconstrained document images is developed using MATLAB R2014b version. Experiments are carried on processor Intel Core i7 with 2.5 GHz frequency and 8GB RAM. Dataset is provided by local software company collaborated with institute for secured application based research. The aim of the collaboration is to develop an application for automatic line segmentation from handwritten text images.

### 5.2 Performance Evaluation

The performance evaluation of the method is based on counting the number of matches between the entities detected by the algorithm and the entities in the ground truth template. A Match Score $(i, j)$ table [12] is employed, representing the matching results of the $j$ ground truth region and the $i$ result region [3]. A match is only considered, if the matching score is equal to or above a specified acceptance threshold $T$. It must be noted that $T$ was set to 95% for the line detection. If $G$ and $F$ are the numbers of ground-truth and result elements, respectively, and $w1, w2,$ and $w3$ are predetermined weights (set to 1, 0.25, 0.25, respectively). The detection rate (DR) is calculated as following.

$$DR = w_1^{og2o_d} + w_2^{og2m_d} + w_3^{mg2o_d}$$

where, $o_g2o_d$ and $o_d2o_g$ denote the one to one matches, $o_g2m_d$ denotes one ground truth to many detected, $m_g2o_d$ denotes many ground truth to one detected, $o_d2m_g$ denotes one detected to many ground truth, and $m_d2o_g$ denotes many detected to one ground truth.

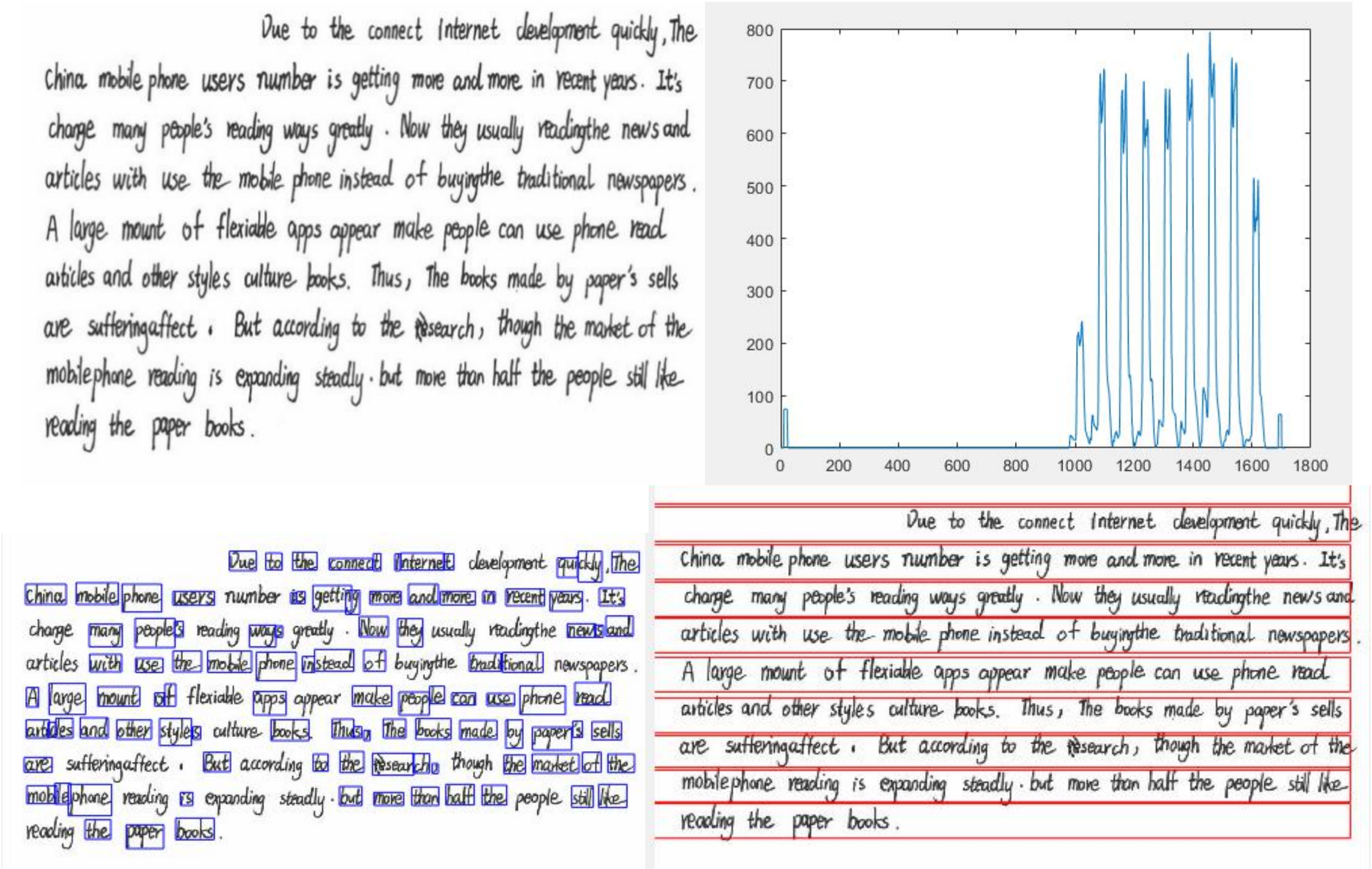

Figure 4: Bounding boxes of significant blobs and red rectangles are final segmented lines on Dataset I

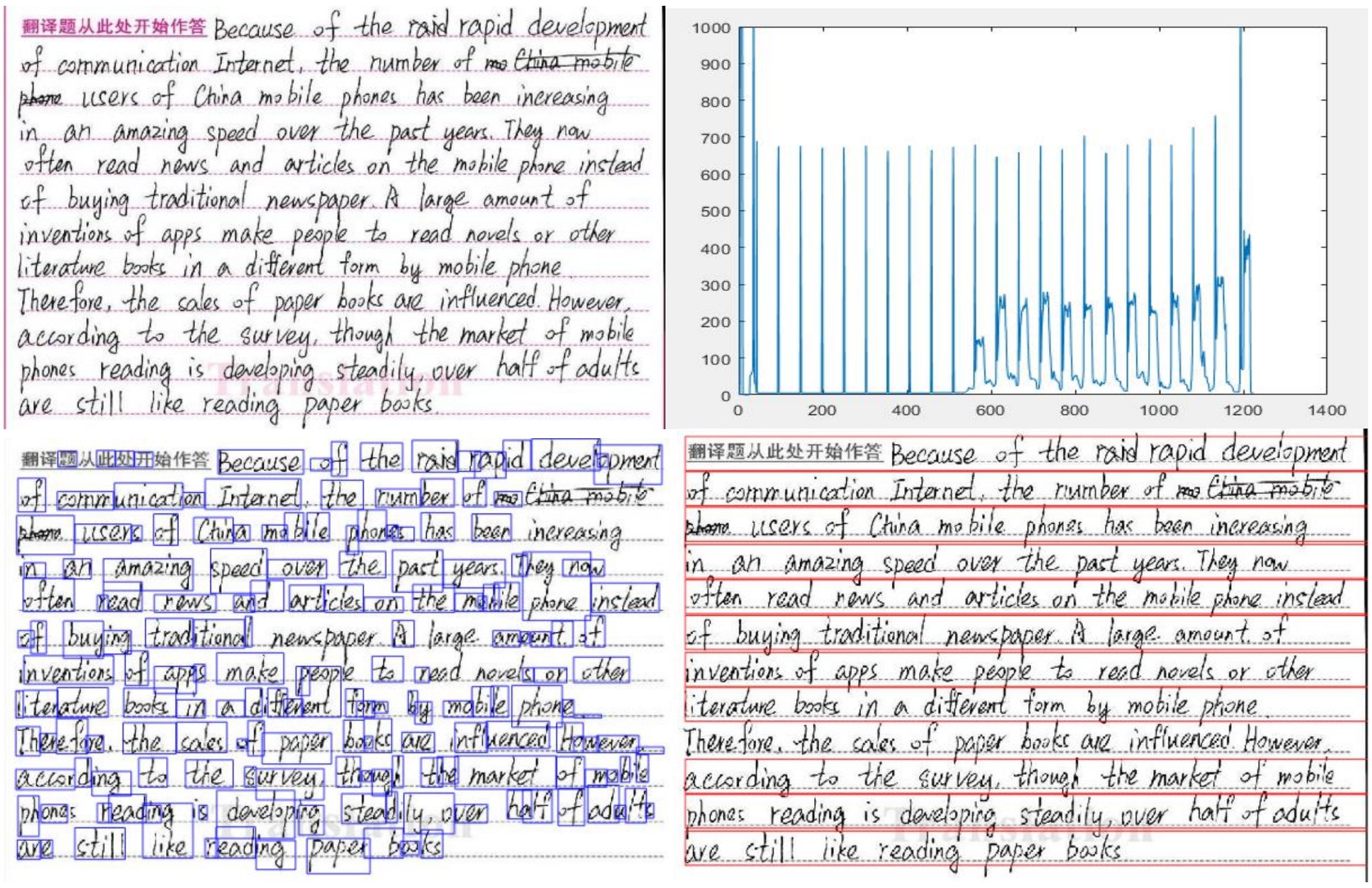

Figure 5: Bounding boxes of significant blobs and red rectangles are final segmented lines on Dataset II

The correct segmentation is calculated by matching the number of segmented lines with the corresponding template as a ground truth text image. If the number of segmented lines from image matches to number of lines present in ground truth text file within the defined threshold, then segmentation is said to be correct, else declare as wrong segmentation.

The results are observed from experiments done on two different types of datasets. Fig. 4 shows the original text image from Dataset I in which base lines are not given in the template. Next plot is a connected component boxes, which boundary coordinates are noted for further matching and alignment. Last plot in Fig 4 is a final output of the proposed segmentation scheme. The proposed scheme is able to segment lines from plain document text images. Similar results are displayed in Fig. 5 while performed on the Dataset II. In this, base lines are available in the templates of ground truth. The proposed method is applied thoroughly and lines are segmented. Observed values are given in Table 1 with total number of images in the dataset and observed detection rates.

Table 1: Detection rates observed on our dataset

| Datasets | Detection rates (in %) |
|---|---|
| Dataset I | 97.58 |
| Dataset II | 98.44 |

## 6. Comparison with related work

The observed values achieved from comparison are given in Table 2 on historical record set of line segmentation from the handwritten images. These techniques are very common and elementary approach for line segmentation task. The Hough transformation based method used an iterative hypothesis-validation strategy. At each stage, the best text-line hypothesis is generated in the Hough domain considering fluctuations of the text-line components. Afterwards, the validity of the line is in using a proximity criterion. Therefore, due to hypothetical analysis, it produces ambiguous components. Projection based method giving 97.31% accuracy, while fuzzy means clustering based technique observe 93% accuracy. In another work, line segmentation using Hough transformation observed 95.67% accuracy. It can be stated clearly that existing techniques does not perform good for line segmentation task. However, the proposed scheme is tested on more than 12,900 document images and it successfully provides a significant range of detection rate on both datasets.

Some of the difficult cases are shown in Fig. 6-8 to justify the performance of the proposed scheme. It is capable to segment lines properly irrespective to the location and size of the words or sentences on the document. That is, unconstrained handwritten lines are also being detected and segmented with minimal error. In Fig. 6, texts are written in unpredictable format and with long wide spaces. On similar observation, in Fig. 7, text seems to be written in circular/round format and in Fig. 8 only three incomplete lines are written in unconstrained manner on entire page. The proposed scheme is capable to segment lines from each of such difficult images efficiently. Thus, the proposed scheme is proven robust, feasible, and highly effective to work on unconstrained datasets.

Table 2: Line segmentation rates given in the literature from conventional methods

| Method | Dataset | Detection rate (in %) |
|---|---|---|
| Projection profile | 720 images | 97.31 |
| Fuzzy based | 1864 images | 93 |
| Hough transformation based | ICDAR 07 | 95.67 |
| Proposed method | Our dataset (13000 images) | 98.01 |

6.1 Cross-Validation on Benchmark Datasets

The proposed scheme is tested on publicly available datasets namely *IAM* and *ICDAR09*, for the uniform evaluation. These both datasets are benchmark and most relevant datasets for text line segmentation. The performance of the proposed method is verified and cross-examined on both the datasets. With this kind of comparison, the uniform evaluation is observed and proposed scheme is made generalized on different kind of datasets.

The obtained result is shown in Fig. 9. It is clearly visible by the results, that the proposed scheme is able to segment lines from document images from *IAM* significantly. In the resultant (Fig. 9), one segmented line is overlapped with two red lines ($m_d2o_g$ error) and hence falls under specified threshold under performance metric. This is an allowed range in the detection rate errors as described in the Section 4.2.

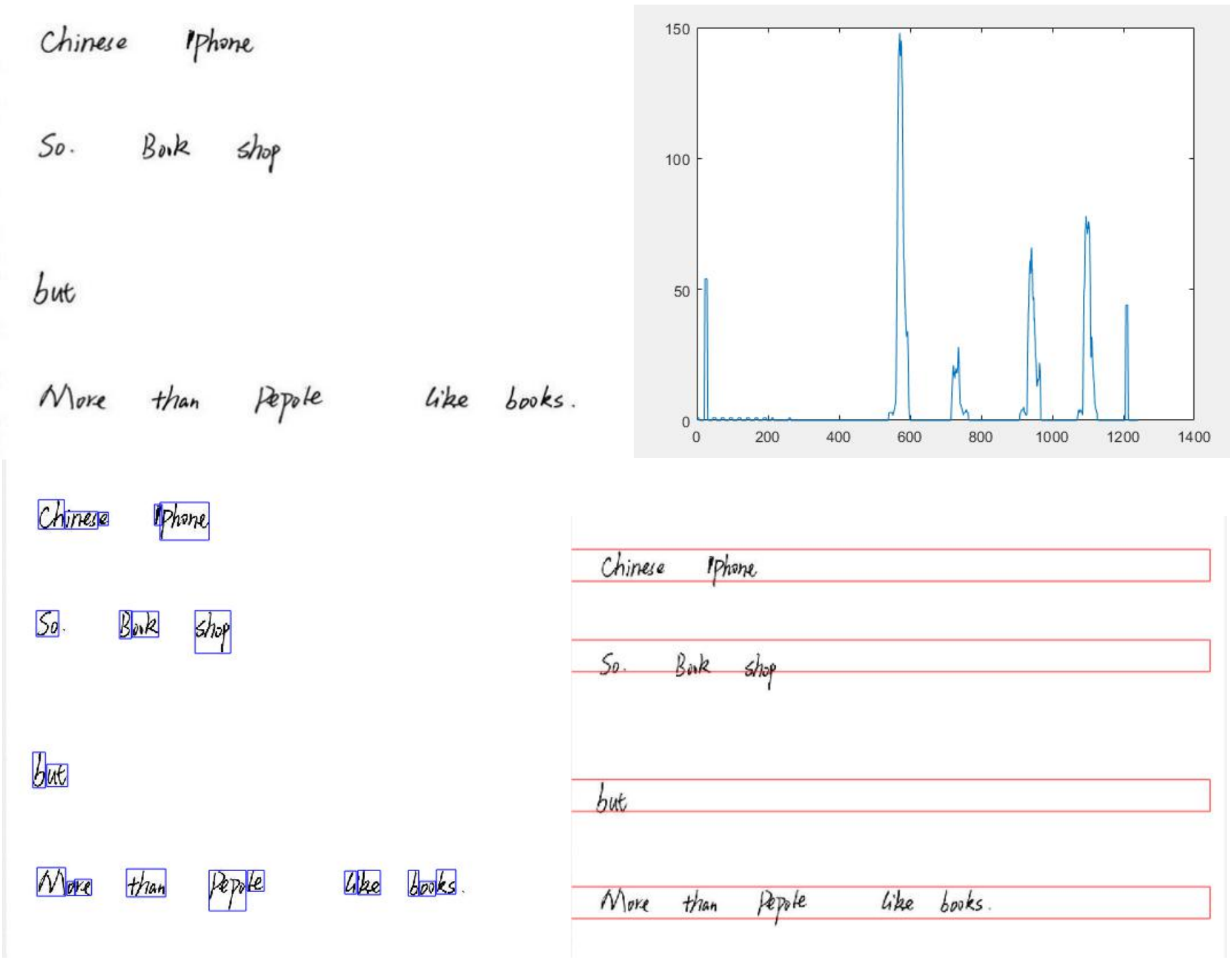

Figure 6: Difficult cases on our datasets and segmented lines as accurate results

Similarly, the result obtained of the proposed scheme on *ICDAR09* dataset is illustrated in Fig. 10. Though the lines are straight segmenting and hence rectangles are drawn, therefore it is leading a kind of error. Error ($m_d2o_g$) is displayed due to many lines are representing to segment one ground truth line, *i.e.*, many to one error type is achieved. But this type of error also belongs to the acceptable range under performance metric, hence accountable for measuring the overall successful detection rate.

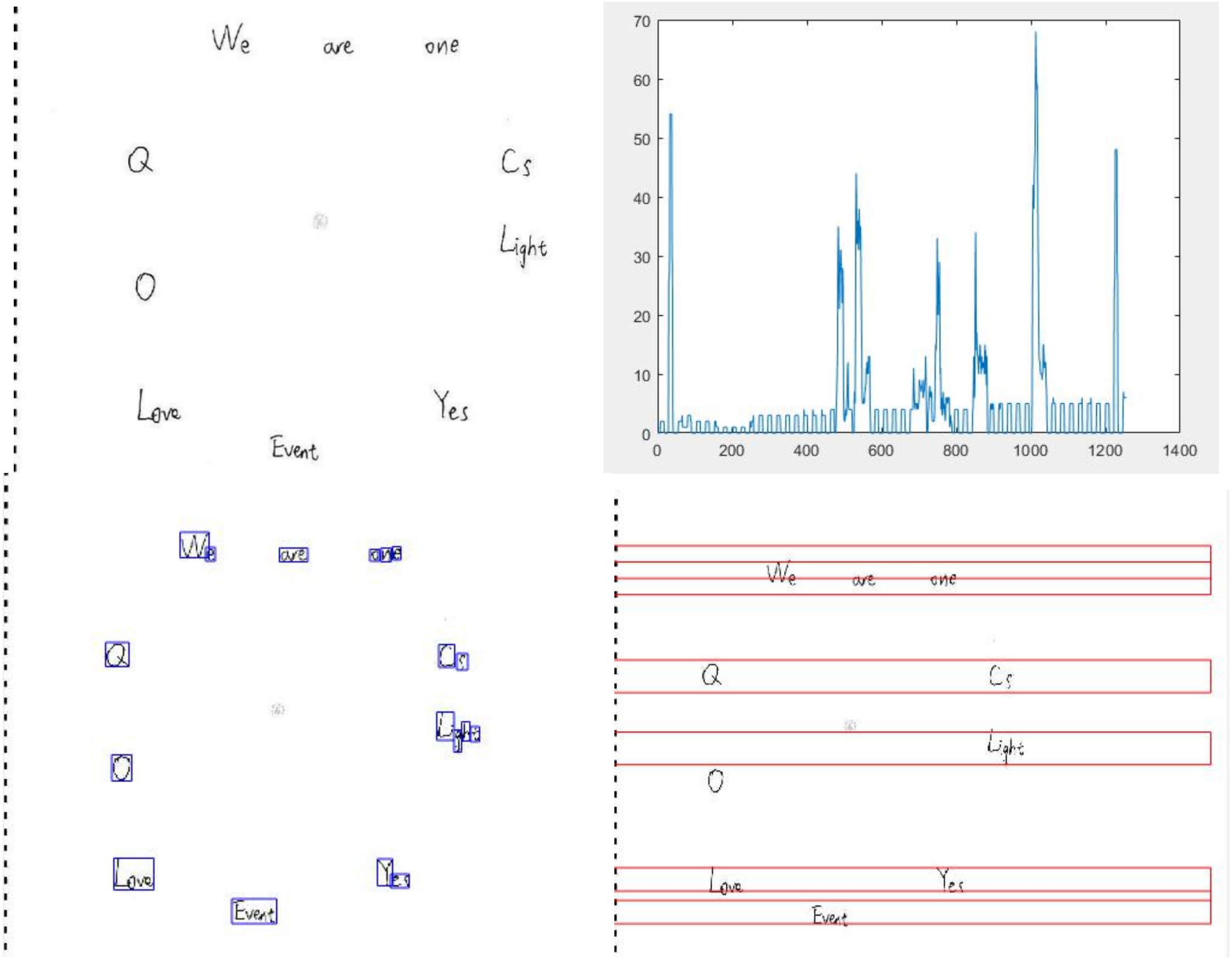

Figure 7: Difficult cases on our datasets and segmented lines

Some of the techniques are thoroughly implemented and tested on our dataset as well on benchmark datasets (Refer Table 3). Further, the pictorial outcomes are given in the Fig. 11, where first column represents the results obtained on our constructed dataset followed by *IAM* and *ICDAR* datasets respectively. The first row shows the results obtained from Alizera [1], which achieved 89.43% on our dataset and 70% and 88.5% detection rates on *IAM* and *ICDAR* datasets respectively. Second row represents the results obtained from Louloudis [3] as only 46.6% detection rate on our dataset while 91.55% and 96% accuracy on *IAM* and *ICDAR* datasets respectively.

The last column shows the outcomes of the different methods on our dataset. Papvasiliou [5] achieved 87.88% detection rate on our dataset, whereas 98.96% on *IAM* and 72.5% on *ICDAR*09 datasets, respectively. The small lines come from the method developed by Papvassilou shows the correct segmented lines, which are to be extended. The actual numbers of lines are counted with extending these lines as it was not required to obtain the count of lines. With the proper analysis of the execution of such algorithms, it can be stated that the proposed method performs better than other methods. Moreover, the system is quite feasible and easy to implement.

Table 3: Comparison of detection rates achieved on different datasets

| | Datasets |
|---|---|

| | IAM | ICDAR | Our Dataset |
|---|---|---|---|
| Papvasiliou [5] | 72.5 | 98.96 | 87.88 |
| Alireza [1] | 70 | 88.5 | 89.43 |
| Louloudis [3] | 91.55 | 96 | 46.6 |
| Proposed Scheme | 91.99 | 96 | 98.01 |

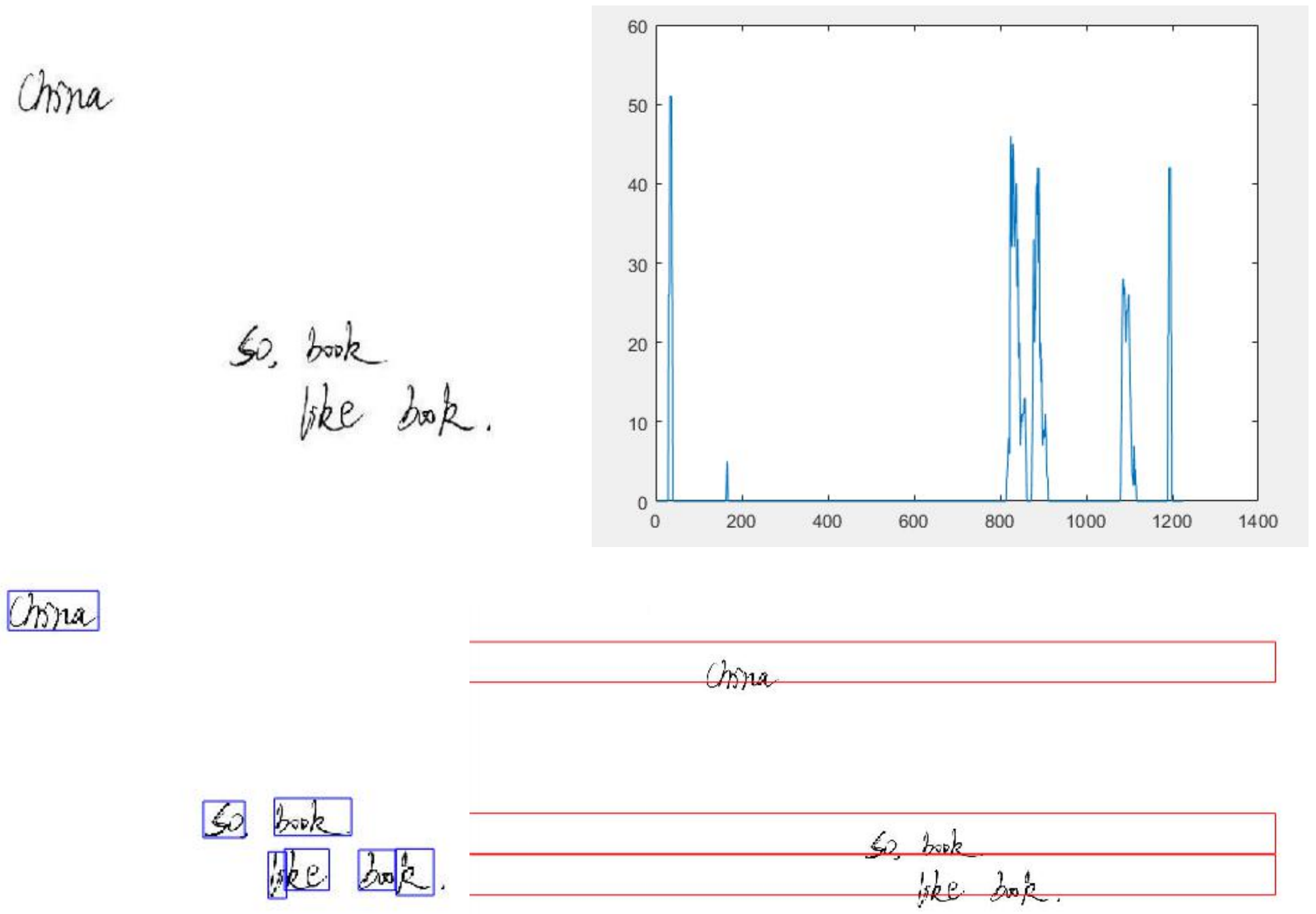

Figure 8: Difficult cases on our datasets and segmented lines

7. Conclusion

The proposed scheme incorporates an adaptive approach to provide feasibility to deal with unconstrained datasets. The method is able to measure text heights with respect to document size. Moreover, the complexity of the method is drastically reduced as compared to various conventional methods. It observes high segmentation accuracy over others. The system is feasible to implement, efficient and less complexed. The system provides 97.58% and 98.44% segmentation accuracy from both the datasets I and II, respectively. Further, it observes 91.99% and 96% detection rates on *IAM* and *ICDAR09* datasets, respectively. The experimental results are cross validated on benchmark datasets to show the efficient and effective solution. Future work extends for including other native language handwritings around the world.

Acknowledgment

Authors would like to thank Dr. Zheng for fruitful discussion and providing dataset from company to test and validate for real-time applications.

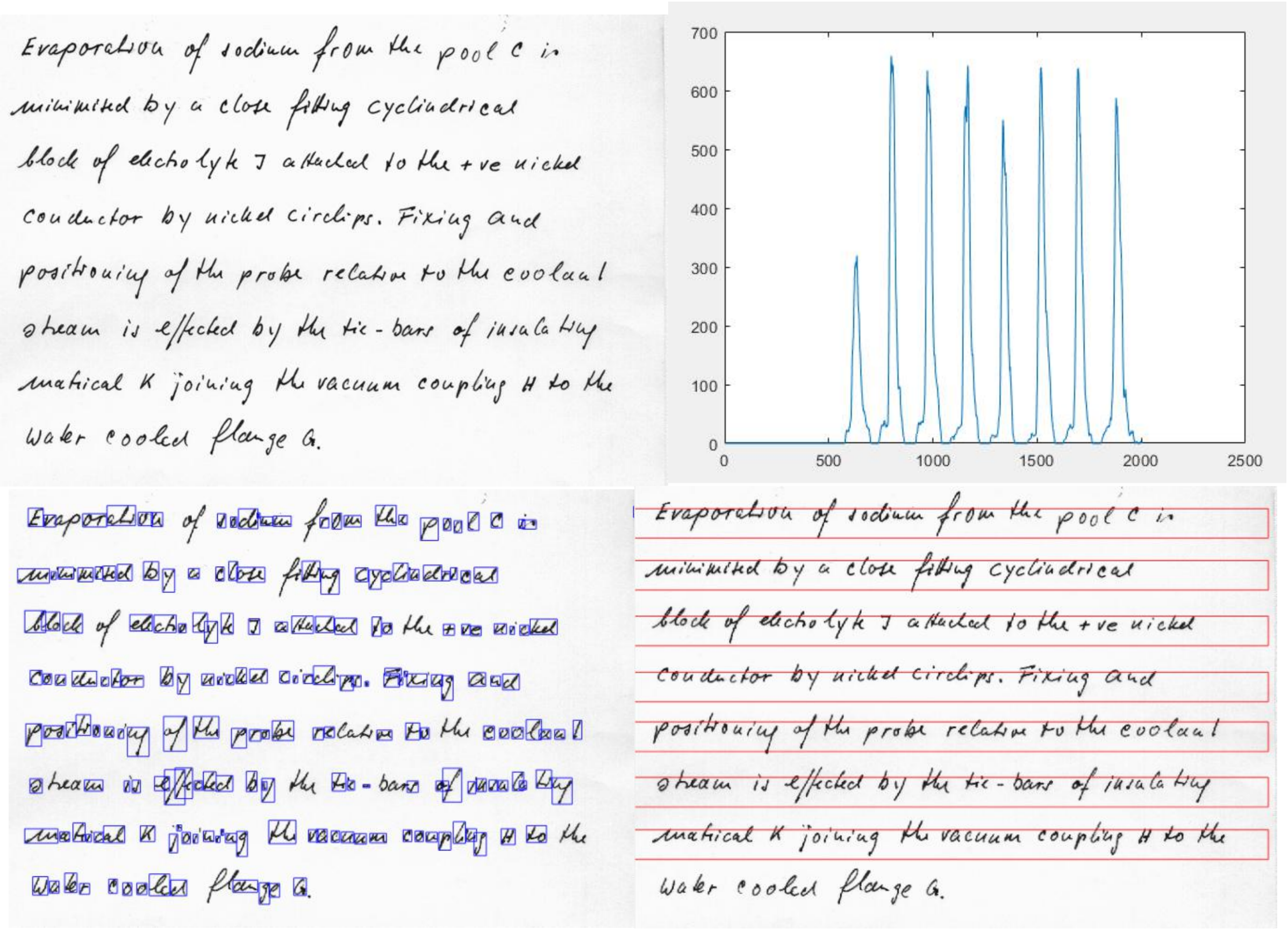

Figure 9: Bounding boxes of significant blobs and red rectangles are final segmented lines on IAM

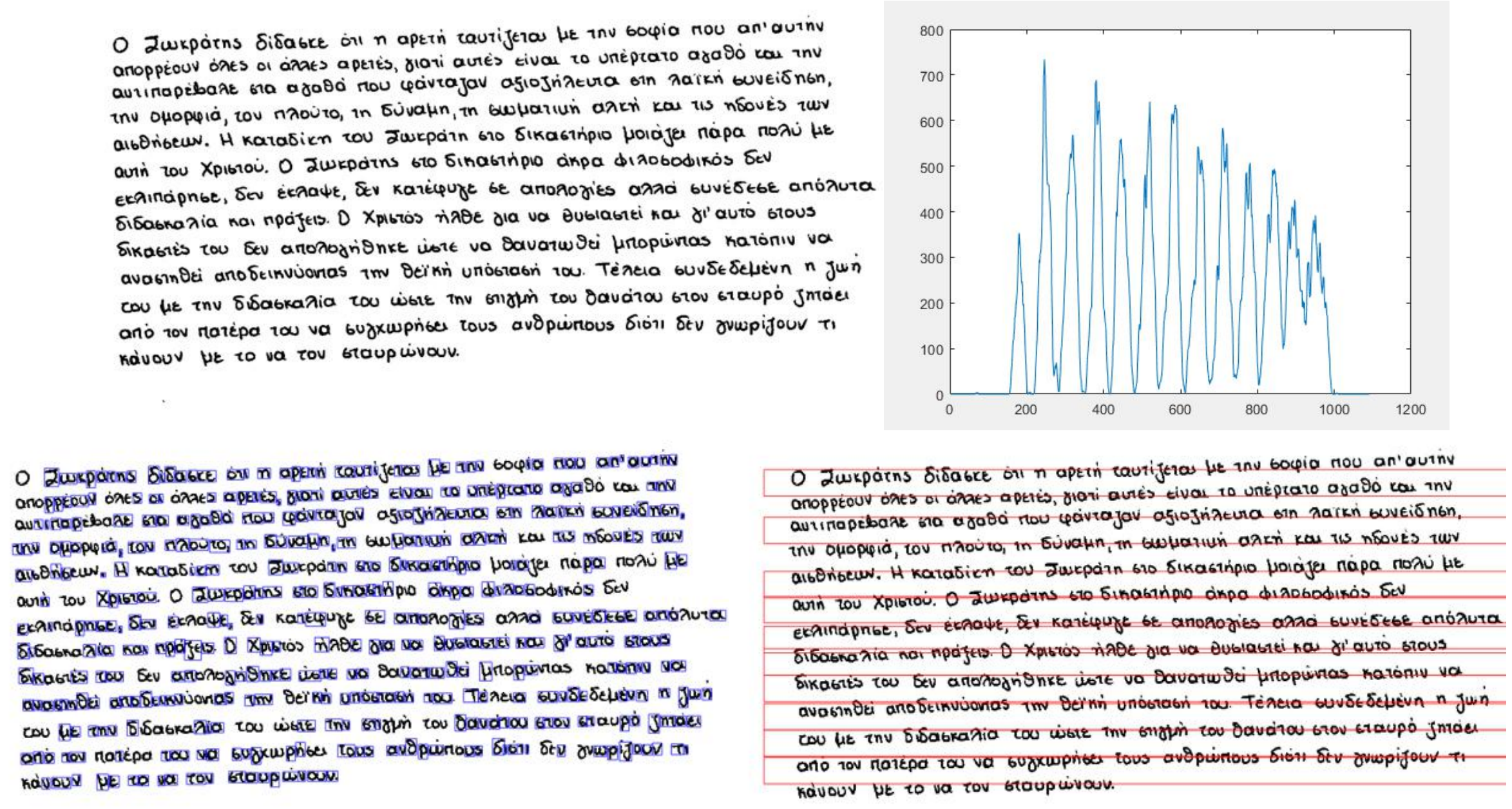

Figure 10: Bounding boxes of significant blobs and red rectangles are final segmented lines on *ICDAR09*

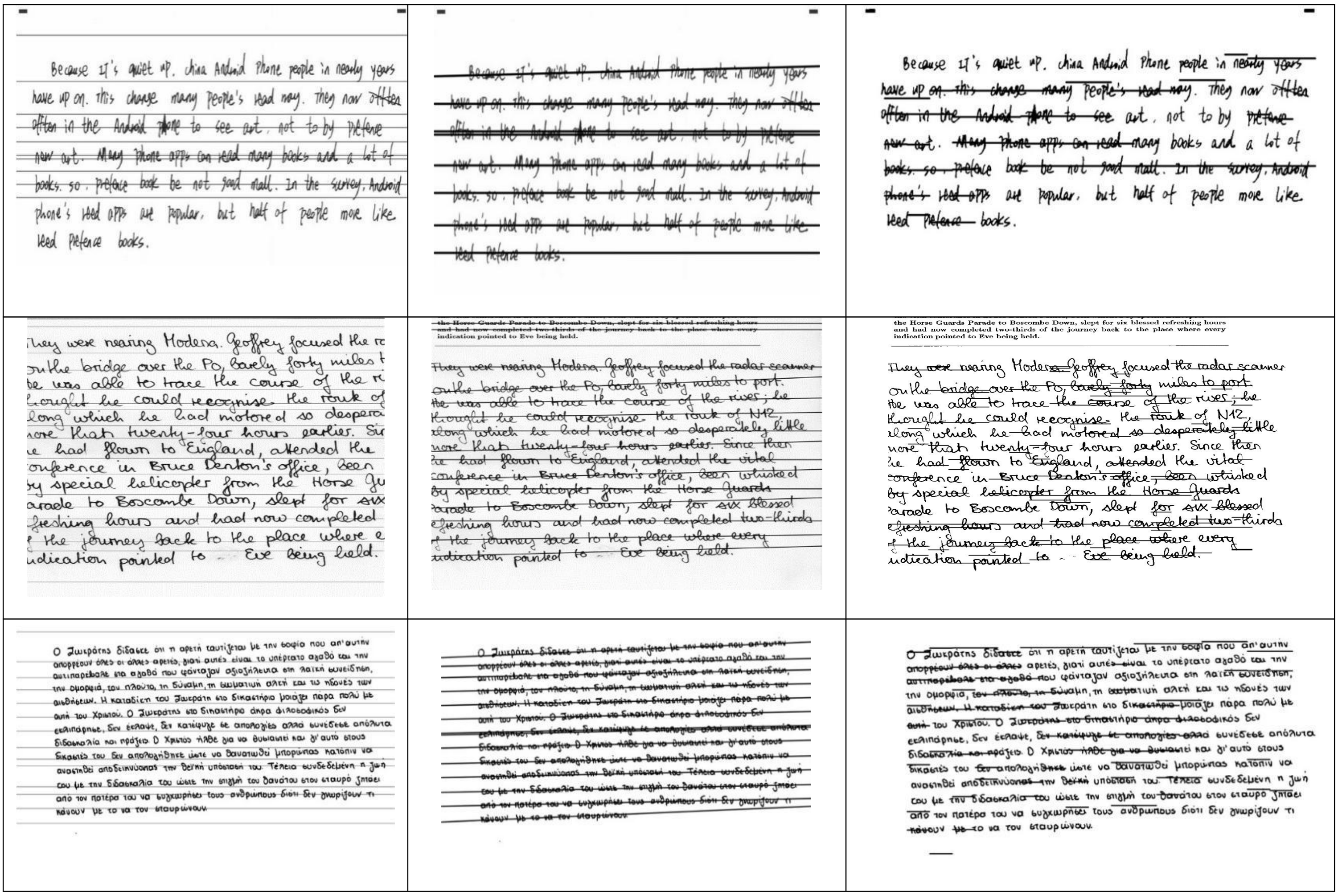

Figure 11: Results obtained from each implemented technique on our dataset as well as on benchmark datasets

Compliance with Ethical Standards

The authors declare that they have no conflict of interest. For this type of study formal consent is not required.